%% file: main.tex
\documentclass[11pt]{article}
\usepackage[preprint]{acl}
\usepackage{times}
\usepackage{latexsym}
\usepackage[T1]{fontenc}
\usepackage[utf8]{inputenc}
\usepackage{microtype}
\usepackage{inconsolata}
\usepackage{graphicx}
\usepackage{amsmath}
\usepackage{booktabs}
\usepackage{caption}
\usepackage{placeins}
\title{Outcome--Mass Alignment: Controlling Length-Dependent Trajectory
Influence in Multi-Turn LLM Agent Reinforcement Learning}
\author{
  \textbf{Dawei Zhang\textsuperscript{1,2}},
  \textbf{Yubin Huangfu\textsuperscript{3}},
  \textbf{Jessica Fan\textsuperscript{4}},
  \textbf{Boyu Wei\textsuperscript{1,2}}
\\
  \textbf{Likang Wu\textsuperscript{5}},
  \textbf{Hongke Zhao\textsuperscript{5}},
  \textbf{Ming He\textsuperscript{3,*}},
  \textbf{Zhu Teng\textsuperscript{6}}
\\[0.4em]
  {\normalfont\small\textsuperscript{1}University of Chinese Academy of Sciences, China}
\\
  {\normalfont\small\textsuperscript{2}Institute of Software, Chinese Academy of Sciences, China}
\\
  {\normalfont\small\textsuperscript{3}AI Lab at Lenovo Research, China}
\\
  {\normalfont\small\textsuperscript{4}University of North Carolina at Chapel Hill, USA}
\\
  {\normalfont\small\textsuperscript{5}Tianjin University, China\quad
  \textsuperscript{6}Beijing Jiaotong University, China}
\\
  {\normalfont\small\textsuperscript{*}Corresponding author:
  \href{mailto:heming01@foxmail.com}{heming01@foxmail.com}}
}
\hypersetup{
  pdftitle={Outcome-Mass Alignment: Controlling Length-Dependent Trajectory Influence in Multi-Turn LLM Agent Reinforcement Learning},
  pdfauthor={Dawei Zhang, Yubin Huangfu, Jessica Fan, Boyu Wei, Likang Wu, Hongke Zhao, Ming He, Zhu Teng},
  pdfsubject={Multi-turn LLM agent reinforcement learning},
  pdfkeywords={LLM agents, multi-turn reinforcement learning, group-relative policy optimization, trajectory reweighting, outcome normalization, loss aggregation}
}

\begin{document}

\maketitle

\begin{abstract}
To keep long interaction contexts tractable, stepwise group-relative training
for multi-turn LLM agents materializes each source trajectory into multiple
response rows and valid tokens. This expansion creates an outer-accounting
problem distinct from within-trajectory credit assignment: row-based outcome
statistics can count one episode outcome multiple times, while token-level
reduction can assign unequal aggregate objective mass across trajectories. We
term these effects \emph{outcome-statistics multiplicity} and
\emph{objective-mass imbalance}. We introduce Outcome--Mass Alignment (OMA), a
two-component method that makes both accounting choices explicit. Outcome
Alignment computes outcome statistics over unique source trajectories, while
Mass Alignment controls aggregate objective mass after within-trajectory
aggregation of optimizer-native token objectives. Neither redesigns the base
optimizer's credit assignment, importance ratios, or clipping.
Across matched ALFWorld GRPO comparisons at two model scales, where failed
trajectories typically expand much more than successful ones, OMA improves
matched-horizon endpoint success by 7.0--9.4 percentage points. Ordinary
WebShop exhibits weak or reversed expansion--outcome coupling and
correspondingly mixed effects, whereas in a success-oriented WebShop setting,
WebShop-SOL, OMA improves success nAUC from 0.399 to 0.511. Component ablations
and GiGPO and GSPO comparisons further support component
separability and cross-optimizer compatibility. These results show that
trajectory expansion always defines an implicit weighting rule, but its
practical effect depends on how expansion couples to outcomes and
optimizer-native updates.
\end{abstract}

\input{sections/01_introduction}

\input{sections/02_preliminaries}

\input{sections/03_outcome_mass_alignment}

\input{sections/04_experiments_analysis}

\input{sections/05_related_work}

\input{sections/06_conclusion}

\paragraph{Use of Generative AI.}
Generative AI tools were used to assist with language editing, manuscript
drafting, and code refinement. The authors reviewed and verified all
AI-assisted outputs and take full responsibility for the technical content,
implementation, experimental results, claims, references, and conclusions.

\label{page:main-content-end}

\input{sections/07_limitations}

\bibliography{references}

\clearpage
\appendix
\raggedbottom
\setcounter{dbltopnumber}{3}
\renewcommand{\dbltopfraction}{0.9}
\renewcommand{\textfraction}{0.1}
\input{sections/appendix_additional_seeds}
\input{sections/appendix_original}

\end{document}

%% file: sections/01_introduction.tex
\section{Introduction}
\label{sec:introduction}

Large language models are increasingly used as agents that perceive, reason,
and act through repeated interaction with external environments. Representative
settings include embodied household tasks in ALFWorld
\citep{shridhar2021alfworld} and goal-directed web interaction in WebShop
\citep{yao2022webshop}. These tasks require long-horizon planning under sparse
feedback: an agent may make tens of decisions before receiving a verifiable
episode-level outcome. Reinforcement learning is therefore a natural
post-training paradigm, and group-relative methods such as GRPO
\citep{shao2024deepseekmath} offer a
scalable, critic-free way to learn from multiple attempts at the same task.
Extending this success to multi-turn agents,
however, requires deciding how a long environment interaction is presented to
the language-model optimizer.

\begin{figure*}[t]
    \centering
    \includegraphics[width=\textwidth]{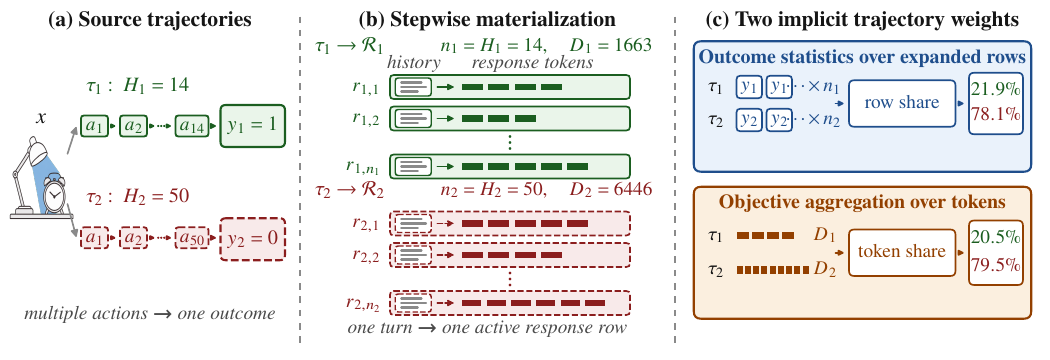}
    \caption{Expansion-induced trajectory reweighting under stepwise
    materialization. (a) Same-task, same-checkpoint ALFWorld rollouts: a
    14-turn success and a 50-turn failure, each with one episode outcome.
    (b) One-response-row-per-turn materialization expands $\tau_i$ into $n_i$
    rows containing $D_i$ valid response tokens. (c) Row-based statistics
    weight trajectory outcomes by $n_i$, whereas token-mean reduction assigns
    objective mass by $D_i$.}
    \label{fig:intro-motivation}
\end{figure*}

Multi-turn pipelines use two main materialization patterns. Trajectory-wise
optimization concatenates all states and model outputs into one sequence, as in
RAGEN \citep{wang2025ragen}, preserving full history but causing the effective
context to grow with the interaction horizon. Stepwise optimization instead
materializes each turn as a response row conditioned on a managed history,
keeping individual contexts tractable but expanding each source trajectory into
a variable number of optimizer-facing units
(Figure~\ref{fig:intro-motivation}(b)).

A direct stepwise adaptation of GRPO broadcasts the same trajectory-level
relative advantage to every row, obscuring which decisions helped or harmed.
GiGPO localizes credit through anchor-state groups \citep{feng2025gigpo}, while
HGPO further organizes steps by historical-context consistency
\citep{he2026hgpo}. These methods address which decisions receive credit within
a trajectory. Stepwise expansion, however, creates a separate outer-accounting
problem: after a source trajectory expands into multiple rows and tokens, how
much total statistical and optimization influence should it receive? OMA
targets this trajectory-level influence rather than redesigning
within-trajectory credit assignment.

Figure~\ref{fig:intro-motivation} illustrates how stepwise materialization can
reweight source trajectories: an ALFWorld task yields a 14-turn success with
1,663 response tokens and a 50-turn failure with
6,446 tokens, both sampled from the same policy checkpoint. Although each
source trajectory yields one outcome, row-based outcome statistics count the
outcomes with different multiplicities, while token-proportional reduction
assigns different aggregate objective mass
(Figure~\ref{fig:intro-motivation}(c)). For this pair, the failed trajectory
therefore accounts for 78.1\% of the row-expanded outcome population and
79.5\% of the token-weighted objective mass. This \emph{expansion-induced
trajectory reweighting} comprises \emph{outcome-statistics multiplicity} and
\emph{objective-mass imbalance}. Stepwise materialization makes trajectory
influence depend implicitly on row and token expansion, without an explicit
trajectory-level weighting choice.


We introduce \emph{Outcome--Mass Alignment} (OMA), an outer accounting
framework that makes these two choices explicit and controllable. Outcome
Alignment constructs the designated outcome statistics over unique source
trajectories before assigning the resulting learning signals to response rows.
Mass Alignment first aggregates the optimizer-native objective contributions
within each source trajectory and then controls its contribution to the outer
objective.
OMA therefore changes neither environment interaction nor the base optimizer's
native within-trajectory credit, importance-ratio, and clipping mechanisms. Its
two components can be enabled independently or jointly and instantiated with
multiple group-relative policy optimizers.

We evaluate OMA on ALFWorld and WebShop with GRPO, together with
GiGPO and GSPO adapter comparisons. ALFWorld exhibits strong success--failure
expansion asymmetry: across two model scales, OMA improves matched-horizon
success by 7.0--9.4 percentage points. Ordinary WebShop exhibits much weaker or
reversed expansion--outcome coupling and correspondingly mixed effects, whereas
a success-oriented WebShop setting, WebShop-SOL, improves success nAUC from
0.399 to 0.511. These results support a condition-dependent account: trajectory
expansion always defines an implicit weighting rule, but its practical effect
depends on how expansion couples to outcomes and optimizer-native updates.

Our contributions are:
\begin{itemize}
    \item We identify expansion-induced trajectory reweighting in stepwise
    group-relative agent training, separating outcome-statistics multiplicity
    from objective-mass imbalance
    (Sections~\ref{sec:trajectory-reweighting}--\ref{sec:mass-alignment}).
    \item We introduce Outcome--Mass Alignment, a two-component framework for
    explicit trajectory-level outcome accounting and objective weighting
    (Sections~\ref{sec:outcome-alignment}--\ref{sec:oma-instantiations}).
    \item We validate this condition-dependent account through component
    ablations on ALFWorld and WebShop and GiGPO and GSPO comparisons
    (Section~\ref{sec:experiments}).
\end{itemize}

%% file: sections/02_preliminaries.tex
\section{Preliminaries}
\label{sec:preliminaries}

\paragraph{Long-horizon agentic tasks.}
Given a task $x$ and an initial condition, training samples a group of $K$
source trajectories $\{\tau_i\}_{i=1}^{K}$, where
$\tau_i=((o_{iu},a_{iu}))_{u=1}^{H_i}$. Here $o_{iu}$ and $a_{iu}$ are the
observation and textual action at turn $u$, and $H_i$ is the realized horizon.
Each trajectory receives one processed episode outcome $y_i$, which may be
unavailable until the interaction terminates.

\paragraph{Stepwise policy optimization.}
Stepwise methods materialize turns as separate response rows conditioned on a
managed history \citep{feng2025gigpo,he2026hgpo}, so a source trajectory may
produce variable numbers of rows and valid tokens. OMA studies this row-expanded
optimization setting.

\paragraph{Group-relative policy optimization.}
GRPO avoids a learned value function by estimating relative advantages from
the outcomes of trajectories sampled for the same task
\citep{shao2024deepseekmath}. For $K>1$, its trajectory-level statistics and
standardized advantage are
\begin{equation}
\begin{aligned}
\mu_\tau&=\frac{1}{K}\sum_{i=1}^{K}y_i,\\
s_\tau^2&=\frac{1}{K-1}\sum_{i=1}^{K}(y_i-\mu_\tau)^2,\\
A_i^{\mathrm{GRPO}}&=\frac{y_i-\mu_\tau}{s_\tau+\epsilon}.
\end{aligned}
\label{eq:grpo-advantage}
\end{equation}
Here $\epsilon>0$ is a numerical stabilizer.

%% file: sections/03_outcome_mass_alignment.tex
\section{Outcome--Mass Alignment}
\label{sec:oma}

\begin{figure*}[t]
    \centering
    \includegraphics[width=\textwidth]{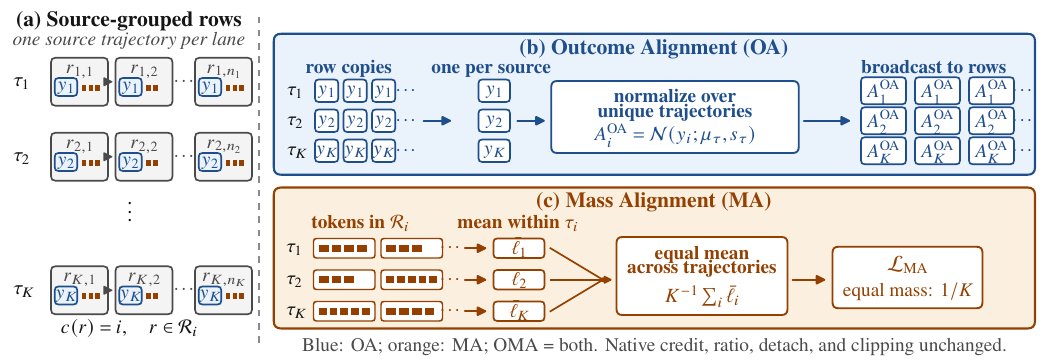}
    \caption{OMA controls outer accounting at the source-trajectory level.
    (a) Rows are grouped by source. (b) OA computes the designated outcome
    component once per unique trajectory and broadcasts it to the corresponding
    rows. (c) MA averages native token objectives within trajectories and then
    across trajectories, assigning each source outer mass $1/K$. Native credit,
    ratios, detach, and clipping remain unchanged.}
    \label{fig:oma-overview}
\end{figure*}

Outcome--Mass Alignment (OMA) is a two-component method for controlling
trajectory-level outer accounting in stepwise group-relative training. For
trajectories expanded into response rows, Outcome Alignment counts each episode
outcome once in the designated group statistics, while Mass Alignment controls
aggregate policy-objective mass across the corresponding rows and tokens. OMA
operates after rollout collection without redesigning within-trajectory credit
assignment. Figure~\ref{fig:oma-overview} summarizes the method.

\subsection{Expansion-Induced Trajectory Reweighting}
\label{sec:trajectory-reweighting}

\paragraph{Stepwise materialization.}
Building on Section~\ref{sec:preliminaries}, let $\mathcal{R}_i$ contain the
active response rows materialized from source trajectory $\tau_i$, and let
$c(r)=i$ map row $r$ back to its source. For token position $p$,
$m_{rp}\in\{0,1\}$ is the valid-token mask. The number of rows, valid tokens in
a row, and total valid-token count of a trajectory are
\begin{equation}
\begin{aligned}
\mathcal{R}_i &= \{r:c(r)=i\}, &
n_i &= |\mathcal{R}_i|,\\
T_r &= \sum_p m_{rp}, &
D_i &= \sum_{r\in\mathcal{R}_i}T_r .
\end{aligned}
\label{eq:oma-setup}
\end{equation}
Thus the training representation follows
$\tau_i\rightarrow\mathcal{R}_i\rightarrow
\{(r,p):r\in\mathcal{R}_i,m_{rp}=1\}$. Turns and rows are distinct objects,
even though our current pipeline emits one active response row per turn.

\paragraph{Native stepwise objective.}
Let $\widetilde A_r$ denote the learning signal produced by the base optimizer
for row $r$. After the optimizer resolves its native advantage, importance
ratio, detach, and clipping operations, let
$\ell_{rp}(\theta;\widetilde A_r)$ be its unreduced token-position objective
contribution, where $\theta$ denotes the policy parameters. We omit its
arguments when unambiguous. For a complete group reduction, common token- and
row-mean objectives are
\begin{equation}
\begin{aligned}
\mathcal{L}_{\mathrm{token}}
&=
\frac{1}{\sum_jD_j}
\sum_i\sum_{r\in\mathcal{R}_i}\sum_p
m_{rp}\ell_{rp},\\
\mathcal{L}_{\mathrm{row}}
&=
\frac{1}{\sum_jn_j}
\sum_i\sum_{r\in\mathcal{R}_i}
\frac{1}{T_r}\sum_p m_{rp}\ell_{rp} .
\end{aligned}
\label{eq:token-mean}
\end{equation}
Consequently, trajectory $i$ receives coefficient mass
$D_i/\sum_jD_j$ under token mean and $n_i/\sum_jn_j$ under row mean.

\paragraph{Outcome-statistics multiplicity.}
If the same episode outcome $y_i$ is attached to every row, row-based
statistics treat it as appearing $n_i$ times. In particular,
\begin{equation}
\mu_{\mathrm{row}}
=
\frac{\sum_i n_i y_i}{\sum_i n_i},
\qquad
\mu_\tau
=
\frac{1}{K}\sum_i y_i .
\label{eq:outcome-populations}
\end{equation}
For example, a successful trajectory with $(y_1,n_1)=(1,1)$ and a failed
trajectory with $(y_2,n_2)=(0,3)$ yield
$\mu_{\mathrm{row}}=0.25$, although the two source trajectories have
$\mu_\tau=0.5$. More generally, under the uniform distribution over
source trajectories,
\begin{equation}
\mu_{\mathrm{row}}-\mu_\tau
=
\frac{\operatorname{Cov}_i(n_i,y_i)}
{\operatorname{E}_i[n_i]} .
\label{eq:outcome-coupling}
\end{equation}
Here $\operatorname{E}_i$ and $\operatorname{Cov}_i$ take the trajectory index
$i$ uniformly over the group.
Variable expansion therefore creates representation sensitivity; a systematic
shift arises when row multiplicity couples to the processed outcome. Row-based
normalization can also change the corresponding variance statistic.

\paragraph{Objective-mass imbalance.}
Even when the advantage is constructed correctly, Eq.~\eqref{eq:token-mean}
shows that the outer reduction still weights a source trajectory according to
its expanded token or row count. Outcome-statistics multiplicity and
objective-mass imbalance therefore occur at different stages and require
different corrections.

\subsection{Outcome Alignment}
\label{sec:outcome-alignment}

Let $\mathcal{N}$ be the base optimizer's episode-normalization map and let
$s_{\mathrm{row}}$ be the scale over the expanded row population, computed
under the same convention as $s_\tau$. A row-expanded implementation uses
$\mathcal{N}(y_{c(r)};\mu_{\mathrm{row}},s_{\mathrm{row}})$; Outcome Alignment
instead uses the unique-trajectory statistics from
Eq.~\eqref{eq:grpo-advantage}:
\begin{equation}
A_i^{\mathrm{OA}}
=\mathcal{N}(y_i;\mu_\tau,s_\tau).
\label{eq:outcome-alignment}
\end{equation}
For standardized GRPO, $\mathcal{N}(y;\mu,s)=(y-\mu)/(s+\epsilon)$, while
mean-only variants retain $y-\mu$. The aligned episode component is then
broadcast to the rows of $\tau_i$. Outcome Alignment changes the population used by
$\mathcal{N}$, not the optimizer's state- or history-conditioned credit
mechanism. Replicating rows with the same trajectory identity and outcome
therefore cannot change the aligned component, and implementations that already
normalize over unique trajectories recover their native computation.

\subsection{Mass Alignment}
\label{sec:mass-alignment}

We choose a source-trajectory-uniform outer objective, assigning equal
coefficient mass to each sampled trajectory regardless of its expanded length.
For a complete group, Mass Alignment realizes this objective by averaging native
token contributions within trajectories and then across trajectories:
\begin{equation}
\mathcal{L}_{\mathrm{MA}}
=
\frac{1}{K}\sum_{i=1}^{K}
\left[
\frac{1}{D_i}
\sum_{r\in\mathcal{R}_i}\sum_p
m_{rp}\ell_{rp}(\theta;\widetilde A_r)
\right].
\label{eq:mass-alignment}
\end{equation}
Each complete source trajectory consequently receives total outer mass $1/K$,
independent of its row or token count. This removes accidental
expansion-proportional weighting.

In practice, rows from one trajectory may be split across actor-loss
reductions. For a reduction containing rows $B$, we retain each trajectory's
fractional mass:
\begin{equation}
\begin{aligned}
&M_B=
\sum_{r\in B}\frac{T_r}{D_{c(r)}},\\
&\mathcal{L}_{\mathrm{MA},B}(\theta;\widetilde A)\\
&\quad=
\frac{1}{M_B}
\sum_{r\in B}\frac{1}{D_{c(r)}}
\sum_p m_{rp}\ell_{rp}(\theta;\widetilde A_r).
\end{aligned}
\label{eq:mass-alignment-minibatch}
\end{equation}
A complete trajectory contributes one unit to $M_B$; a partially present
trajectory contributes the fraction of its valid tokens contained in $B$.
Exact $1/K$ outer mass and representation invariance hold for complete-group
reductions; separately normalized partial reductions are not invariant to
arbitrary repartitioning.

The effect of native token reduction is explicit in a complete batch. Let
\[
\bar{\ell}_i
=
\frac{1}{D_i}
\sum_{r\in\mathcal{R}_i}\sum_p
m_{rp}\ell_{rp} .
\]
Then
\begin{equation}
\mathcal{L}_{\mathrm{token}}
-\mathcal{L}_{\mathrm{MA}}
=
\frac{\operatorname{Cov}_i(D_i,\bar{\ell}_i)}
{\operatorname{E}_i[D_i]} .
\label{eq:mass-coupling}
\end{equation}
Unequal token counts always define an implicit weighting rule, while their
update-level effect depends on how expanded token count couples to the
optimizer-native per-trajectory objective contribution.

\subsection{Unified OMA and Optimizer Instantiations}
\label{sec:oma-instantiations}

The components can be enabled independently. OA-only replaces the designated
episode-level signal while retaining the native reducer; MA-only retains the
native signal $\widetilde A_r$ and applies
Eq.~\eqref{eq:mass-alignment-minibatch}. Full OMA applies both corrections:
\begin{equation}
\mathcal{L}_{\mathrm{OMA},B}(\theta)
=
\mathcal{L}_{\mathrm{MA},B}
\bigl(\theta;\widetilde A^{\mathrm{OA}}\bigr),
\label{eq:unified-oma}
\end{equation}
\noindent where $\widetilde A^{\mathrm{OA}}$ denotes the base optimizer's
native row signals after replacing only its episode-outcome component.

OMA therefore requires access to that component and to unreduced objective
contributions after native credit, ratio, detach, and clipping. GRPO broadcasts
the aligned episode advantage; GiGPO retains anchor grouping, step credit, and
mixing while replacing only its episode term; GSPO retains sequence ratios,
detach, and clipping before Mass Alignment. Table~\ref{tab:supp-oma-instantiations} summarizes
these interfaces.

%% file: sections/04_experiments_analysis.tex
\section{Experiments and Analysis}
\label{sec:experiments}

We evaluate OMA in target and boundary regimes. ALFWorld provides strong
success--failure expansion asymmetry and supports component and dynamics
analyses; ordinary WebShop tests weak or reversed coupling, WebShop-SOL modifies
the training outcome, and GiGPO and GSPO probe cross-optimizer compatibility.

\subsection{Experimental Setup and Expansion--Outcome Diagnosis}
\label{sec:experimental-setup}

We evaluate Qwen2.5-1.5B/3B-Instruct \citep{qwen2024qwen25} on ALFWorld
\citep{shridhar2021alfworld} and WebShop \citep{yao2022webshop}. ALFWorld reports
validation success; WebShop reports exact success and official task score. GRPO
is primary, with GiGPO and GSPO compatibility tests. In our stepwise
implementation, Base uses row-expanded outcome statistics and token-mean loss
reduction. OA changes the outcome-statistics population, MA changes the outer
reduction, and OMA applies both.

We additionally evaluate OMA in WebShop-SOL, a success-oriented training setting
that modifies the training outcome and uses successful-only termination.
Evaluation retains the standard WebShop metrics. Let
$z_i$ be exact success, $q_i$ the best trajectory score, and
$N_i^{\mathrm{att}}$/$N_i^{\mathrm{fail}}$ purchase/failed-purchase counts:
\begin{equation}
y_i^{\mathrm{SOL}} =
\begin{cases}
10-0.1N_i^{\mathrm{fail}}, & z_i=1,\\
-0.05, & z_i=0 \text{ and } N_i^{\mathrm{att}}=0,\\
q_i-0.1N_i^{\mathrm{fail}}, & z_i=0 \text{ and } N_i^{\mathrm{att}}>0.
\end{cases}
\label{eq:webshop-sol}
\end{equation}

The original main-text comparisons use the first 150 updates, eight trajectories
per task group, batch size 16, one PPO epoch, seed 0, and four GPUs. We report endpoints and
nAUC, with early $[0,50]$, middle $[50,110]$, and late $[110,150]$ windows;
full hyperparameters, hardware, and diagnostic definitions are supplementary.
ALFWorld failures commonly reach the 50-turn limit while successes terminate
earlier; WebShop lengths are closer and can reverse. This diagnoses
outcome--expansion coupling, but not the mass--gradient covariance governing MA,
which is assessed through ablations and frozen-batch geometry.

\subsection{Main Results}
\label{sec:main-results}

\input{generated/main_alfworld}
\begin{figure*}[t]
    \centering
    \includegraphics[width=\textwidth]{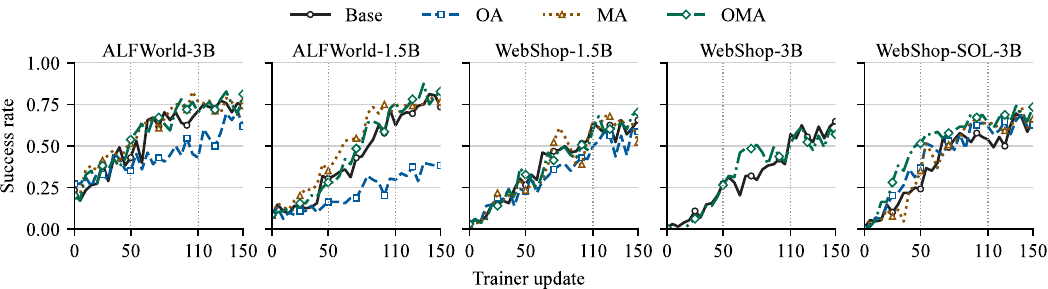}
    \caption{Matched-horizon GRPO success curves on ALFWorld, WebShop, and
    WebShop-SOL. Dotted lines mark stage boundaries at updates 50 and 110;
    Table~\ref{tab:main-webshop} reports score metrics.}
    \label{fig:performance-curves}
\end{figure*}

\input{generated/main_webshop}

Tables~\ref{tab:main-alfworld} and \ref{tab:main-webshop} show
condition-dependent effects. OMA raises ALFWorld success from $0.734$ to
$0.828$ at 1.5B and from $0.742$ to $0.812$ at 3B; the GiGPO and
GSPO adapters are also positive. Ordinary WebShop is mixed or reversed, whereas
WebShop-SOL improves from $0.648$ to $0.734$ endpoint success and from $0.399$
to $0.511$ success nAUC. Overall, these results show that systematic differences
in row and token expansion between high- and low-outcome trajectories can make
implicit trajectory reweighting consequential, and that OMA mitigates this
effect through explicit trajectory-level accounting.
Figure~\ref{fig:performance-curves} gives the matched GRPO curves used in the
stage and component analyses.
Additional two-seed experiments on ALFWorld-3B and WebShop-1.5B are reported in
Appendix~\ref{sec:additional-seeds}, covering Base/OA/MA/OMA and Base/OMA,
respectively. Their mean results favor OMA over Base on ALFWorld-3B, while
WebShop-1.5B shows higher endpoints but lower nAUC.

\subsection{Outcome and Mass Alignment Ablations}
\label{sec:oma-ablations}

OA corrects the outcome population, MA corrects outer objective mass, and OMA
is invariant to both representation changes under the fixed-batch accounting
contract. Tables~\ref{tab:main-webshop} and
\ref{tab:ablation-alfworld} report the completed four-way matrices.

The ALFWorld ablations show that the corrections are not interchangeable. OA
removes row-population sensitivity but leaves token-weighted mass and
underperforms Base. MA obtains the strongest overall success nAUC
($0.601$ at 3B and $0.510$ at 1.5B). At 3B it leads the early and middle
windows ($0.388$ and $0.682$), while OMA leads late ($0.754$) and gives the
strongest endpoints at both scales ($0.812$ and $0.828$). At 1.5B, MA likewise
leads early and middle ($0.213$ and $0.594$), while OMA leads late ($0.776$).
Ordinary WebShop has no uniformly dominant component. On WebShop-SOL, OA
improves success nAUC, MA leads score nAUC, and OMA leads strict success and the
score endpoint. Full window-level results are supplementary.

\input{generated/ablation_alfworld}

\subsection{Gradient Geometry under Expansion Asymmetry}
\label{sec:gradient-geometry}

\input{generated/gradient_geometry_main}

Table~\ref{tab:gradient-geometry-main} summarizes five representative GRPO
batches from a ratio-one frozen-batch audit. We recombine the same
per-trajectory full-model gradient basis under Raw, OA, MA, and OMA. The
near-symmetric WebShop batch remains close to Raw, whereas stronger ALFWorld
expansion asymmetry is associated with larger norm changes, especially for MA,
and negative OA--MA cosine in several batches. The negative cosine shows that
the two corrections can act in opposing directions rather than as a scalar
rescaling. Because old log probability is the detached current value, clipping
is inactive and no optimizer step is taken; the audit isolates outer-accounting
effects. Fixed-basis tests confirm OA invariance to row duplication, MA
invariance to token-mass inflation, and OMA invariance to both within the audited
contract. Table~\ref{tab:gradient-geometry-full} provides the complete matrix and cross-policy
rows; additional negative-control checks are described in the supplement.

\subsection{Learning Dynamics and Mechanism Analysis}
\label{sec:learning-dynamics}

Figure~\ref{fig:supp-dynamics-curves} reports entropy, KL, episode-length, and signed-PG
traces for the completed ALFWorld-3B and WebShop-SOL ablations. They exhibit
method-associated differences consistent with the stage- and metric-dependent
component results. Detailed interpretation is supplementary.

\subsection{Cross-Optimizer Composability}
\label{sec:optimizer-composability}

The GiGPO Base--OMA pair improves endpoint success from $0.883$ to $0.898$ and
nAUC from $0.631$ to $0.673$; the GSPO pair improves them from $0.656$ to
$0.703$ and from $0.428$ to $0.492$. Together with GRPO, these results support
compatibility between outer accounting and different native
objectives.

\subsection{WebShop as a Boundary Case}
\label{sec:webshop-boundary}

Ordinary WebShop shows the mixed behavior expected under weak or reversed
expansion--outcome asymmetry. At 1.5B, endpoint success improves from $0.656$ to
$0.703$ and task score from $0.762$ to $0.788$, but success nAUC changes from
$0.391$ to $0.390$ and score nAUC from $0.633$ to $0.579$. At 3B, success
reverses from $0.648$ to $0.570$ and score from $0.814$ to $0.714$. Thus changing
trajectory accounting has a smaller and less consistent effect when longer
trajectories are not systematically associated with failure.

WebShop-SOL increases the contrast for exact success while retaining graded
failures. OMA then improves success endpoint and nAUC, MA leads score nAUC, and
OMA leads the score endpoint. Together with the distinct dynamics, this
within-setting ablation is consistent with outcome geometry modulating the
component effects. More generally, larger accounting effects are expected when
row multiplicity couples to outcome statistics or expanded token count couples
to per-trajectory objective gradients.

Taken together, the results suggest a practical diagnosis. Outcome--expansion
coupling determines whether row replication changes outcome statistics.
Expanded token count determines native coefficient mass, while coupling to
per-trajectory gradients determines the update-level effect of that weighting.
The row-population effect motivates OA; the mass-weighting effect motivates MA
and is better assessed through component ablations or frozen-batch geometry
than through trajectory length alone.
Stronger coupling on both axes is associated with larger accounting effects,
whereas weak or reversed coupling is associated with mixed component rankings.

We treat ordinary WebShop as the boundary case and SOL as a modified-outcome
ablation. Because model scale and termination also differ, their contrast is
not an isolated causal test.

\FloatBarrier

%% file: generated/main_alfworld.tex
\begin{table*}[t]
    \centering
    \begin{tabular}{@{}lrrrrrrrr@{}}
        \toprule
        Method & Pk & Lk & Cl & Ht & Co & P2 & All & nAUC \\
        \midrule
        \multicolumn{9}{@{}l}{\textit{Qwen2.5-1.5B-Instruct}} \\
        GRPO & 86.7 & 57.1 & 81.2 & 58.3 & 65.0 & \textbf{70.0} & 73.4 & 43.5 \\
        \textbf{GRPO + OMA} & \textbf{87.5} & \textbf{75.0} & \textbf{100.0} & \textbf{83.3} & \textbf{80.0} & 69.6 & \textbf{82.8} & \textbf{46.5} \\
        \addlinespace[1pt]
        \multicolumn{9}{@{}l}{\textit{Qwen2.5-3B-Instruct}} \\
        GRPO & 94.6 & \textbf{77.8} & 72.0 & 46.2 & 57.9 & 72.0 & 74.2 & 55.5 \\
        \textbf{GRPO + OMA} & 96.3 & 71.4 & 93.3 & 57.1 & 73.7 & 76.5 & 81.2 & 58.2 \\
        GiGPO & 93.3 & 71.4 & 90.6 & \textbf{100.0} & 75.0 & \textbf{95.0} & 88.3 & 63.1 \\
        \textbf{GiGPO + OMA} & \textbf{100.0} & 76.2 & \textbf{100.0} & 71.4 & \textbf{84.2} & 94.1 & \textbf{89.8} & \textbf{67.3} \\
        GSPO & 80.0 & 71.4 & 65.6 & 50.0 & 60.0 & 55.0 & 65.6 & 42.8 \\
        \textbf{GSPO + OMA} & 96.3 & 61.9 & 80.0 & 57.1 & 52.6 & 52.9 & 70.3 & 49.2 \\
        \bottomrule
    \end{tabular}
    \caption{ALFWorld main results (\%). Pk/Lk/Cl/Ht/Co/P2 denote Pick, Look, Clean, Heat, Cool, and Pick2; All is endpoint success.}
    \label{tab:main-alfworld}
\end{table*}

%% file: generated/main_webshop.tex
\begin{center}
    \begin{minipage}{\columnwidth}
    \centering
    \setlength{\tabcolsep}{3pt}
    \begin{tabular}{@{}lrrrr@{}}
        \toprule
        Method & \multicolumn{2}{c}{Score} & \multicolumn{2}{c}{Success} \\
        \cmidrule(lr){2-3}\cmidrule(lr){4-5}
         & Endpoint & nAUC & Endpoint & nAUC \\
        \midrule
        \multicolumn{5}{@{}l}{\textit{WebShop -- Qwen2.5-1.5B-Instruct}} \\
        Base & 76.2 & \textbf{63.3} & 65.6 & 39.1 \\
        OA & \textbf{79.7} & 63.1 & 58.6 & 33.5 \\
        MA & 68.2 & 58.3 & 52.3 & \textbf{39.2} \\
        OMA & 78.8 & 57.9 & \textbf{70.3} & 39.0 \\
        \addlinespace[1pt]
        \multicolumn{5}{@{}l}{\textit{WebShop -- Qwen2.5-3B-Instruct}} \\
        Base & \textbf{81.4} & \textbf{51.8} & \textbf{64.8} & 32.4 \\
        OMA & 71.4 & 51.8 & 57.0 & \textbf{34.4} \\
        \addlinespace[1pt]
        \multicolumn{5}{@{}l}{\textit{WebShop-SOL -- Qwen2.5-3B-Instruct}} \\
        Base & 83.1 & 68.0 & 64.8 & 39.9 \\
        OA & 83.1 & 67.7 & 62.5 & 44.0 \\
        MA & 83.8 & \textbf{70.3} & 66.4 & 42.5 \\
        OMA & \textbf{85.6} & 69.4 & \textbf{73.4} & \textbf{51.1} \\
        \bottomrule
    \end{tabular}
    \captionof{table}{Combined WebShop and WebShop-SOL GRPO main and component results (\%). Each metric reports its endpoint and nAUC through update 150. WebShop-3B includes Base and OMA only.}
    \label{tab:main-webshop}
    \end{minipage}
\end{center}

%% file: generated/ablation_alfworld.tex
\begin{center}
    \begin{minipage}{\columnwidth}
    \centering
    \setlength{\tabcolsep}{5pt}
    \begin{tabular}{@{}llrr@{}}
        \toprule
        Setting & Variant & Succ. & nAUC \\
        \midrule
        ALFWorld (3B) & Base & 74.2 & 55.5 \\
         & OA & 61.7 & 44.8 \\
         & MA & 77.3 & \textbf{60.1} \\
         & OMA & \textbf{81.2} & 58.2 \\
        \addlinespace[1pt]
        ALFWorld (1.5B) & Base & 73.4 & 43.5 \\
         & OA & 38.3 & 22.3 \\
         & MA & 76.6 & \textbf{51.0} \\
         & OMA & \textbf{82.8} & 46.5 \\
        \bottomrule
    \end{tabular}
    \captionof{table}{ALFWorld GRPO component ablations (\%). Succ. reports endpoint success.}
    \label{tab:ablation-alfworld}
    \end{minipage}
\end{center}

%% file: generated/gradient_geometry_main.tex
\begin{table}[!ht]
    \centering
    \small
    \begin{tabular}{lrrrrr}
        \toprule
        Env. & $r_{\mathrm{tok}}$ & OA & MA & OMA & $\cos(\mathrm{OA},\mathrm{MA})$ \\
        \midrule
        WS & 0.99 & 0.94 & 1.02 & 0.96 & 0.996 \\
        ALF & 1.14 & 1.19 & 1.26 & 0.94 & 0.261 \\
        ALF & 2.17 & 1.93 & 2.72 & 1.30 & -0.421 \\
        ALF & 3.59 & 2.41 & 3.46 & 1.31 & -0.506 \\
        ALF & 8.78 & 2.44 & 5.05 & 1.60 & -0.368 \\
        \bottomrule
    \end{tabular}
    \caption{Frozen-batch full-model gradient geometry. $r_{\mathrm{tok}}$ is the ratio of mean failed-trajectory to successful-trajectory response tokens. OA, MA, and OMA report gradient-norm ratios to Raw; WS and ALF denote WebShop and ALFWorld.}
    \label{tab:gradient-geometry-main}
\end{table}

%% file: sections/05_related_work.tex
\section{Related Work}
\label{sec:related-work}

\textbf{Policy optimization and objective aggregation.}
Language-model policy optimization builds on PPO
\citep{schulman2017ppo} and critic-free estimators such as RLOO
\citep{ahmadian2024rloo}. GRPO applies prompt-group relative advantages through
a clipped token surrogate \citep{shao2024deepseekmath}; DAPO modifies sampling,
clipping, and token aggregation \citep{yu2025dapo}; Dr.\ GRPO removes
length and within-question variance scaling \citep{liu2025understanding}; and
GSPO moves ratios and clipping to the sequence level \citep{zheng2025gspo}.
Related work balances token- and sequence-level contributions
\citep{zeng2026balanced}, develops length-aware normalization or token
preferences \citep{he2025vlnorm,wang2025lambdagrpo}, and aligns off-policy
optimization with turn structure \citep{li2026sorl}; broader comparisons study
interactions among estimators, environments, and rewards
\citep{wang2025practitionerguide}. These methods modify normalization, credit,
gradient balance, or clipping within optimizer-facing units. OMA instead uses
source identity to define the outcome population and total outer objective mass
after row expansion.

\textbf{Credit assignment in multi-turn agents.}
Multi-turn agents interleave reasoning and environment actions
\citep{yao2023react,abdulhai2023lmrlgym}. Training systems use hierarchical
optimization, curricula, or critic-based step feedback
\citep{zhou2024archer,qi2025webrl,zhou2025sweetrl} within the broader temporal
credit-assignment problem \citep{pignatelli2024credit,zhang2026agenticcredit}.
Turn rewards densify sparse feedback \citep{wei2025turnreward}; hindsight and
hierarchical methods estimate step values or separate planning from execution
\citep{tan2026hcapo,peng2026hiper}. Group-relative methods construct
state- or history-conditioned comparison groups
\citep{feng2025gigpo,he2026hgpo}, or assign and propagate step- or turn-level
credit
\citep{li2026salt,qi2026stapo,wang2026steppo,zong2026at2po}. RAGEN optimizes
whole trajectories \citep{wang2025ragen}, whereas Agent Lightning materializes
executions as training transitions \citep{luo2025agentlightning}. These methods
determine which decisions receive credit or how interactions are materialized;
OMA instead controls each source trajectory's aggregate influence while
preserving native credit, ratios, and clipping.

\textbf{Trajectory-length weighting.}
AsyncWebRL is the closest comparison on trajectory-length weighting
\citep{bai2026asyncwebrl}. In multi-step GRPO, it replaces $1/H_i$ with a
trajectory-independent factor, making total weight proportional to $H_i$ and
strengthening negative gradients from long failures to encourage efficient
termination. OMA neither prefers shorter trajectories nor scales mass with
$H_i$: each source receives equal outer objective mass regardless of its row or
token count. OMA also constructs outcome statistics over unique source
identities before broadcasting them to rows, removing representation
sensitivity. Both methods expose a length-dependent outer-mass design axis but
choose different weighting policies and accounting boundaries.

%% file: sections/06_conclusion.tex
\section{Conclusion}
\label{sec:conclusion}

Stepwise multi-turn training expands one outcome into variable numbers of rows
and tokens, inducing outcome-statistics multiplicity and objective-mass
imbalance. OMA controls both through unique-trajectory statistics and equal
outer trajectory mass while preserving native credit, ratios, and clipping.

Results support a condition-dependent effect: OMA is most beneficial when
expansion couples strongly to outcomes and objective mass, as in ALFWorld,
where MA is strongest early--middle and full OMA late and at the endpoint.
Ordinary WebShop provides a weak or reversed boundary, while WebShop-SOL shows
sensitivity to outcome geometry. GiGPO and GSPO results suggest cross-optimizer
compatibility. Thus expansion defines an implicit trajectory-weighting policy.

%% file: sections/07_limitations.tex
\section*{Limitations}
\label{sec:limitations}

The original main-text experiments use one seed, two environments, and
Qwen2.5 models at 1.5B/3B. Appendix~\ref{sec:additional-seeds} adds two-seed
experiments for ALFWorld-3B and WebShop-1.5B; these summaries provide limited
evidence of run-to-run variation rather than population-level estimates. Matching trainer updates does not equalize generated tokens,
environment interactions, wall-clock time, or compute; mixed GPU generations
also preclude efficiency claims.

Expansion--outcome coupling is policy-conditioned and can change during
training. The reported dynamics are diagnostic rather than causal, and the
ordinary-WebShop/WebShop-SOL contrast also changes model scale and termination.
The gradient audit uses selected ratio-one frozen batches rather than bitwise
historical minibatch replay. OMA requires reliable trajectory identities and
valid-token accounting; its update-local normalization is not invariant to
arbitrary separately normalized repartitioning.

%% file: sections/appendix_additional_seeds.tex
\section{Additional Two-Seed Experiments}
\label{sec:additional-seeds}

We report 12 additional GRPO runs: Base, OA, MA, and OMA on
ALFWorld with Qwen2.5-3B-Instruct (eight runs), and Base and OMA on
WebShop with Qwen2.5-1.5B-Instruct (four runs). Each method uses two distinct
training seeds, displayed as Seed 1 and Seed 2 within its setting. These
labels identify independent runs rather than resumed tracking segments.
The experiment provenance retains the original random-seed values and run
identifiers. The additional runs are summarized separately from the original
main-text experiments.

\begin{figure*}[t]
    \centering
    \includegraphics[width=\textwidth]{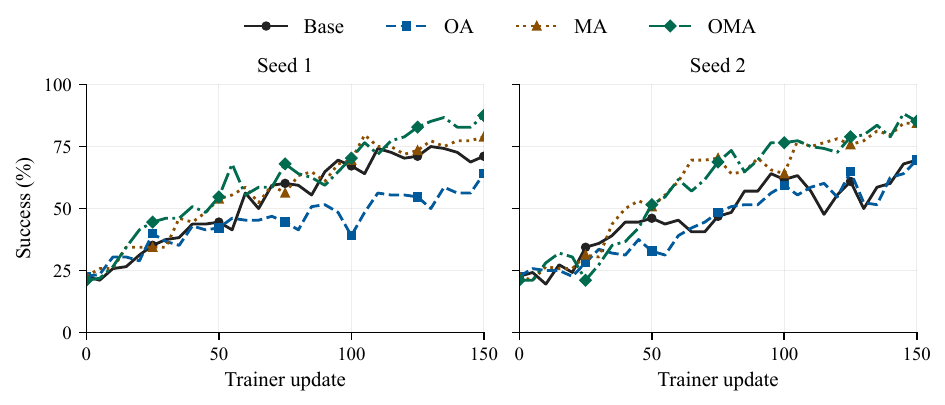}
    \caption{Additional ALFWorld-3B GRPO learning curves. Each panel shows
    one independent training seed for Base, OA, MA, and OMA. Lines connect
    every observed validation checkpoint without smoothing; marker shapes
    and line styles distinguish methods.}
    \label{fig:additional-alfworld}
\end{figure*}

\begin{figure*}[t]
    \centering
    \includegraphics[width=\textwidth]{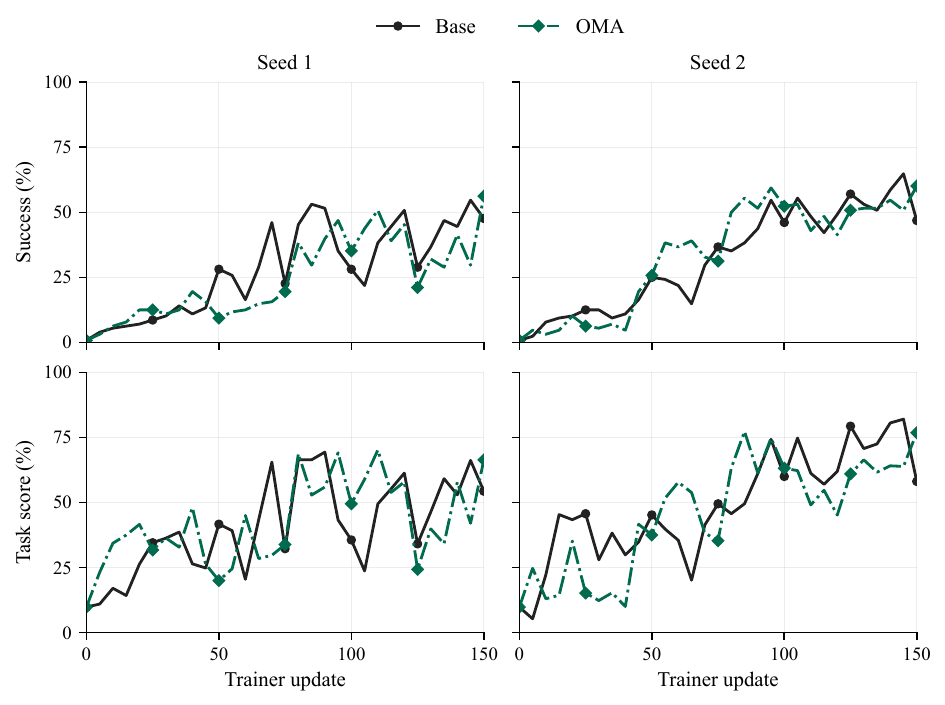}
    \caption{Additional WebShop-1.5B GRPO learning curves. Columns identify
    the two independent training seeds; rows report exact success and
    standard task score. Lines connect all observed checkpoints without
    smoothing. Endpoint gains need not imply higher nAUC.}
    \label{fig:additional-webshop}
\end{figure*}

\paragraph{Protocol and statistics.}
All additional runs use four GPUs, a task-group batch size of 16, and an actor
learning rate of $10^{-6}$. Training seeds vary across the two runs; the
evaluation environment seed remains fixed at 1000. Validation uses greedy
decoding (temperature 0). Each curve contains validation at update 0 and every
five updates through update 150.
We report the observed update-150 endpoint and nAUC over $[0,150]$, using
linear interpolation between observed checkpoints and trapezoidal
integration, as in Appendix~\ref{sec:supp-protocol}.
Tables~\ref{tab:additional-alfworld} and \ref{tab:additional-webshop} show
the per-seed results and arithmetic means with sample standard deviations
across the two seeds ($n=2$, denominator $n-1$). All values are expressed
as percentages. Figures~\ref{fig:additional-alfworld} and
\ref{fig:additional-webshop} show the unsmoothed validation curves.

\input{generated/additional_alfworld}
\input{generated/additional_webshop}

\paragraph{ALFWorld-3B.}
OMA improves endpoint success over Base for both seeds, with the mean
increasing from $70.31\%$ to $86.33\%$; mean success nAUC increases from
$50.56\%$ to $59.60\%$. MA also improves both metrics over Base for both
seeds, whereas OA has lower nAUC for both seeds and a lower or equal endpoint.
OMA has the highest endpoint for each seed. Its mean nAUC exceeds MA's
($59.60\%$ versus $58.26\%$), although their nAUC ranking differs across
the two seeds.

\paragraph{WebShop-1.5B.}
OMA improves endpoint success and task score over Base for both seeds.
Mean success increases from $47.27\%$ to $58.20\%$, and mean task score
from $56.23\%$ to $71.62\%$. The nAUC comparison remains mixed across
seeds: mean success nAUC decreases from $30.29\%$ to $29.15\%$, and
mean score nAUC from $45.37\%$ to $43.89\%$. The endpoint improvements
therefore coexist with less consistent learning-curve improvements.

\paragraph{Evidence scope.}
These experiments provide evidence from two valid training seeds for the
selected configurations. Their means and standard deviations are descriptive.
Runs identified as affected by experiment bugs are excluded, as are the additional
ALFWorld-1.5B and all 7B runs; the original archives are retained for
traceability.

%% file: generated/additional_alfworld.tex
\begin{table*}[t]
\centering
\small
\begin{tabular}{@{}lrrrrrr@{}}
\toprule
& \multicolumn{2}{c}{Seed 1} & \multicolumn{2}{c}{Seed 2} & \multicolumn{2}{c}{Mean $\pm$ SD} \\
\cmidrule(lr){2-3}\cmidrule(lr){4-5}\cmidrule(lr){6-7}
Method & Endpoint & nAUC & Endpoint & nAUC & Endpoint & nAUC \\
\midrule
Base & 71.09 & 54.04 & 69.53 & 47.08 & $70.31 \pm 1.10$ & $50.56 \pm 4.92$ \\
OA & 64.06 & 44.86 & 69.53 & 44.64 & $66.80 \pm 3.87$ & $44.75 \pm 0.16$ \\
MA & 78.91 & 57.45 & 84.38 & 59.08 & $81.64 \pm 3.87$ & $58.26 \pm 1.15$ \\
OMA & 87.50 & 60.79 & 85.16 & 58.41 & $86.33 \pm 1.66$ & $59.60 \pm 1.68$ \\
\bottomrule
\end{tabular}
\caption{Additional ALFWorld-3B GRPO success results (\%). Endpoint is validation success at update 150; nAUC integrates updates 0--150. SD is the sample standard deviation across the two runs. Seed 1 and Seed 2 are presentation labels.}
\label{tab:additional-alfworld}
\end{table*}

%% file: generated/additional_webshop.tex
\begin{table*}[t]
\centering
\small
\begin{tabular}{@{}llrrrr@{}}
\toprule
& & \multicolumn{2}{c}{Score} & \multicolumn{2}{c}{Success} \\
\cmidrule(lr){3-4}\cmidrule(lr){5-6}
Method & Run / summary & Endpoint & nAUC & Endpoint & nAUC \\
\midrule
Base & Seed 1 & 54.35 & 41.10 & 47.66 & 28.41 \\
Base & Seed 2 & 58.11 & 49.63 & 46.88 & 32.17 \\
Base & Mean $\pm$ SD & $56.23 \pm 2.66$ & $45.37 \pm 6.03$ & $47.27 \pm 0.55$ & $30.29 \pm 2.66$ \\
\midrule
OMA & Seed 1 & 66.39 & 42.22 & 56.25 & 24.49 \\
OMA & Seed 2 & 76.84 & 45.57 & 60.16 & 33.80 \\
OMA & Mean $\pm$ SD & $71.62 \pm 7.39$ & $43.89 \pm 2.37$ & $58.20 \pm 2.76$ & $29.15 \pm 6.58$ \\
\bottomrule
\end{tabular}
\caption{Additional WebShop-1.5B GRPO results (\%). Score is the standard task score; Success is exact success. Endpoints and nAUC use the same update-150 horizon as Table~\ref{tab:additional-alfworld}. SD is the sample standard deviation across the two runs.}
\label{tab:additional-webshop}
\end{table*}

%% file: sections/appendix_original.tex
\section{Additional Method Details and Formal Properties}
\label{sec:supp-formal-scope}

This section expands the accounting identities and representation-invariance
claims underlying Outcome--Mass Alignment (OMA).

\paragraph{Scope and assumptions.}
The algebra in the main paper assumes that every active response row retains a
reliable source-trajectory identity, that the episode outcome attached to rows
of one source is identical, and that valid-token masks define the objective
support. Exact equal mass and representation invariance are statements about a
complete group reduction. Equation~\eqref{eq:mass-alignment-minibatch} in the main paper gives the
fractional-mass implementation used when a trajectory is split across actor
microbatches, but arbitrary independently normalized repartitioning is outside
the invariance claim. OMA also assumes that the base optimizer exposes its
designated episode-level outcome component and unreduced native objective after
credit assignment, importance ratios, detach operations, and clipping.

\paragraph{Outcome-statistics identity.}
Let the source index be uniform over the $K$ trajectories. Expanding
$\operatorname{Cov}(n,y)$ gives
\[
\begin{aligned}
\operatorname{Cov}_i(n_i,y_i)
&=
\frac{1}{K}\sum_i n_i y_i
\\
&\quad -
\left(\frac{1}{K}\sum_i n_i\right)
\left(\frac{1}{K}\sum_i y_i\right).
\end{aligned}
\]
Dividing by $\operatorname{E}_i[n_i]$ yields
$\mu_{\mathrm{row}}-\mu_\tau$, proving the identity in main-paper
Eq.~\eqref{eq:outcome-coupling}. Thus row replication changes the processed outcome population exactly
when expansion multiplicity covaries with outcome.

\paragraph{Objective-mass identity.}
Writing the native token mean as
$\sum_i D_i\bar{\ell}_i/\sum_jD_j$ and Mass Alignment as
$K^{-1}\sum_i\bar{\ell}_i$, the same covariance expansion gives
main-paper Eq.~\eqref{eq:mass-coupling}:
\[
\mathcal{L}_{\mathrm{token}}-\mathcal{L}_{\mathrm{MA}}
=
\frac{\operatorname{Cov}_i(D_i,\bar{\ell}_i)}
{\operatorname{E}_i[D_i]}.
\]

\paragraph{Representation-invariance claims.}
Outcome Alignment depends on the set of unique pairs $(i,y_i)$, so duplicating
rows while preserving source identities and outcomes leaves its statistics
unchanged. Mass Alignment first sums valid-token contributions and divides by
the corresponding source total $D_i$; duplicating a representation together
with its token mass multiplies numerator and denominator by the same factor.
Full OMA inherits both properties. The frozen-batch negative controls in
Section~\ref{sec:supp-audit-provenance} verify that each component moves when its
protected accounting rule is deliberately removed.

\paragraph{Nature of the theoretical contribution.}
These results formalize outer-accounting identities and invariance contracts.
The experiments test the predicted representation sensitivity in selected
agent-training regimes, while ordinary WebShop supplies a boundary case.

\section{Complete Experimental Setup}
\label{sec:supp-protocol}

The original main-text comparisons use the first 150 trainer updates
($T_{\mathrm{upd}}=150$). Each update samples eight trajectories per task group
with batch size 16. ALFWorld permits at most 50 environment turns and WebShop at
most 15. The actor learning rate is $10^{-6}$, with one PPO epoch, KL-loss
coefficient $0.01$, and entropy coefficient $0.001$. Validation is performed
before training and every five updates. These original runs use seed 0 and four GPUs. The additional two-seed
experiments are specified separately in Appendix~\ref{sec:additional-seeds}. Some comparisons span A800 and H20 hosts; trainer updates are therefore
the primary protocol, while wall-clock and GPU-hour comparisons remain
observational.

\subsection{Data, Splits, and Preprocessing}
\label{sec:supp-data-metrics}

ALFWorld and WebShop are existing, publicly available benchmarks cited in the
main paper. ALFWorld is used through its text environment and in-distribution
evaluation split. WebShop uses the public
text-shopping environment and its official task goals. WebShop-SOL changes only
the training outcome and successful-only termination rule; it uses the same
WebShop tasks and ordinary WebShop validation metrics.

The training input files are Parquet task descriptors consumed by the common
agent-training loader and bound to the corresponding environment at launch. No
learned data filter, private annotation set, or result-dependent example
selection is used. Benchmark assets remain governed by their original
distributions.

\subsection{Metrics and Their Motivation}

ALFWorld exact success is the fraction of validation tasks completed
successfully. WebShop exact success is the fraction of tasks for which the
requested item and options are purchased; official task score retains graded
partial credit. Exact success is the strict task-completion measure shared
across the two environments, whereas WebShop score exposes progress that binary
success omits.

For a validation metric $g(t)$ at trainer update $t$, we report the endpoint
$g(T_{\mathrm{upd}})$ and normalized area under the curve
\begin{equation}
\operatorname{nAUC}_{[t_0,t_1]}(g)
= \frac{1}{t_1-t_0}\int_{t_0}^{t_1} g(t)\,dt,
\label{eq:supp-nauc}
\end{equation}
using linear interpolation between validation checkpoints. Overall nAUC uses
$[0,T_{\mathrm{upd}}]$; stage analysis uses early $[0,50]$, middle $[50,110]$,
and late $[110,T_{\mathrm{upd}}]$ windows. A run that has not reached the end of
a window is marked pending rather than extrapolated.

Endpoint values measure final matched-horizon quality. Normalized AUC measures
learning over the entire matched update budget and prevents a late checkpoint
from hiding early or middle behavior. Task-family success in ALFWorld diagnoses
whether aggregate gains are concentrated in one interaction type. Actor
entropy, KL loss, episode length, and signed PG loss are secondary mechanism
diagnostics.

We separately inspect actor entropy, actor KL loss, mean episode length, and the
signed policy-gradient (PG) surrogate loss over the full training trajectory.
We do not compute AUC over them. The analysis instead considers their slopes,
extrema, sustained changes, and temporal relationship to validation performance.
Because OA and MA
alter the population or reducer used by the actor objective, the logged entropy,
KL, and PG values are method-native, objective-weighted diagnostics rather than
reducer-neutral policy statistics.

\subsection{Run Count, Lineage, and Randomness}
\label{sec:supp-runs}

Every original main-text method--environment--model cell is computed from one independent
training lineage with seed 0. A lineage may contain multiple physical tracking
segments when training resumes after interruption; resumed segments continue
the same checkpoint namespace and are not additional seeds or independent
replicates. Overlapping steps are resolved in canonical lineage order.

For the original main-text runs, the environment seed is set to 0 for training and offset by 1000 for validation
environment construction. The WebShop environment applies the seed to Python,
NumPy, and PyTorch sampling; ALFWorld receives the seed through its environment
builder. Distributed kernels and asynchronous environment execution are not
claimed to be bitwise deterministic. The original main-text results therefore report one observed optimization
trajectory per cell; the additional experiments in
Appendix~\ref{sec:additional-seeds} report two seeds per included cell.

\paragraph{Paper-facing run manifest.}
ALFWorld includes GRPO at 1.5B and 3B with Base, OA, MA, and OMA, plus
3B GiGPO and GSPO Base--OMA pairs. WebShop includes the four GRPO variants at
1.5B and a Base--OMA pair at 3B. WebShop-SOL includes the four GRPO variants at
3B. Every original cell listed here uses one independent seed-0 lineage; the
additional runs in Appendix~\ref{sec:additional-seeds} are summarized separately.

\subsection{Statistical and Development Scope}
\label{sec:supp-statistical-scope}

The original main-text comparisons report one seed per cell.
Appendix~\ref{sec:additional-seeds} reports per-seed results and two-seed means
and sample standard deviations for the additional runs. No confidence interval
or significance test is reported. Numerical differences are descriptive
comparisons at a matched trainer-update horizon. Final configurations are
recorded below.

\subsection{Final Hyperparameters}
\label{sec:supp-hyperparameters}

The tables below report the final hyperparameter settings used in the original
main-text comparisons, grouping identical values across settings.

\begin{table*}[t]
    \centering
    \small
    \setlength{\tabcolsep}{5pt}
    \begin{tabular}{@{}p{0.32\textwidth}p{0.62\textwidth}@{}}
        \toprule
        Shared setting & Final value \\
        \midrule
        Actor optimization & learning rate $10^{-6}$; weight decay $0.01$; gradient clip $1.0$; one PPO epoch \\
        Standard policy objective & ratio clip $0.2$; low-variance KL loss coefficient $0.01$; entropy coefficient $0.001$ \\
        Reward and advantage controls & no KL term in reward; invalid-action penalty enabled with coefficient $0.1$; standard GRPO normalization by group standard deviation \\
        Training sample unit & 16 task groups per update; 8 trajectories per group \\
        Generation & temperature $1.0$ for training and $0.4$ for validation; top-$p=1.0$; top-$k$ disabled \\
        Model execution & rollout dtype bfloat16; LoRA disabled \\
        Validation and randomness & validation before training and every 5 updates; seed $0$; one node with 4 GPUs \\
        \bottomrule
    \end{tabular}
    \par\smallskip
    \textit{Shared settings; ``top-$k$ disabled'' denotes top-$k=-1$.}
    \par
    \vspace{4pt}
    \setlength{\tabcolsep}{4pt}
    \begin{tabular}{@{}p{0.22\textwidth}p{0.25\textwidth}p{0.22\textwidth}p{0.22\textwidth}@{}}
        \toprule
        Setting & ALFWorld & WebShop & WebShop-SOL \\
        \midrule
        Model & Qwen2.5-1.5B/3B-Instruct & Qwen2.5-1.5B/3B-Instruct & Qwen2.5-3B-Instruct \\
        Prompt / response limit & $2048$ / $512$ & $4096$ / $512$ & $4096$ / $512$ \\
        History / max turns & $2$ / $50$ & $2$ / $15$ & $2$ / $15$ \\
        Validation batch & $128$ & $128$ & $128$ \\
        PPO minibatch / per-GPU microbatch & $256$ / $32$ & $64$ / $8$ & $64$ / $8$ \\
        Tensor parallelism & $1$ at 1.5B; $2$ at 3B & $1$ at 1.5B; $2$ at 3B & $2$ \\
        \bottomrule
    \end{tabular}
    \caption*{Final scientifically relevant hyperparameters: shared settings
    followed by environment and model deltas.}
    \label{tab:supp-final-hparams}
\end{table*}

\paragraph{Accounting variants.}
Base uses the native row-expanded signal and token mean. OA replaces only the
outcome statistics with the unique-trajectory population. MA retains the native
signal but reduces the valid-token objective within each trajectory and then
across trajectories. OMA applies both changes while preserving the base
optimizer's ratio, detach construction, and clipping.

\paragraph{Optimizer-specific settings.}
GRPO uses $\gamma=\lambda=1$ and the standard clipped tokenwise objective.
GiGPO uses $\gamma=0.95$, $\lambda=1$, mean/std step normalization, and
step-advantage weight $1$. GSPO uses $\gamma=\lambda=1$ and its sequence loss
with lower and upper clips $0.0003$ and $0.0004$. Their OMA variants replace the
designated episode component and outer reducer only. WebShop-SOL uses
successful-only termination, success reward $10$, failed-purchase penalty
$0.1$, and no-purchase penalty $0.05$.

\subsection{Infrastructure and Software}
\label{sec:supp-infrastructure}

The original main-text experiments use one node and four GPUs. Recorded run metadata identify two
accelerator classes: NVIDIA A800 PCIe GPUs with 80 GB each and NVIDIA H20 GPUs
with 96 GB each. Recorded host profiles contain 1007--1510 GB system memory and
Intel Xeon Gold 6330 or Intel Xeon Platinum 8458P CPUs. Runs span Ubuntu
22.04.5/24.04.2, Python 3.10.20/3.12.11/3.12.13, and CUDA 12.4/12.9.

The two principal software families are PyTorch 2.8.0, Transformers 4.57.3,
vLLM 0.11.0, and Ray 2.49.2/2.50.0 for the later ALFWorld and WebShop runs; and
PyTorch 2.6.0, Transformers 4.51.1, vLLM 0.8.2, and Ray 2.49.2 for WebShop-SOL
and earlier WebShop runs.

\input{generated/supplement_stage_results}

\begin{figure*}[t]
    \centering
    \includegraphics[width=\textwidth]{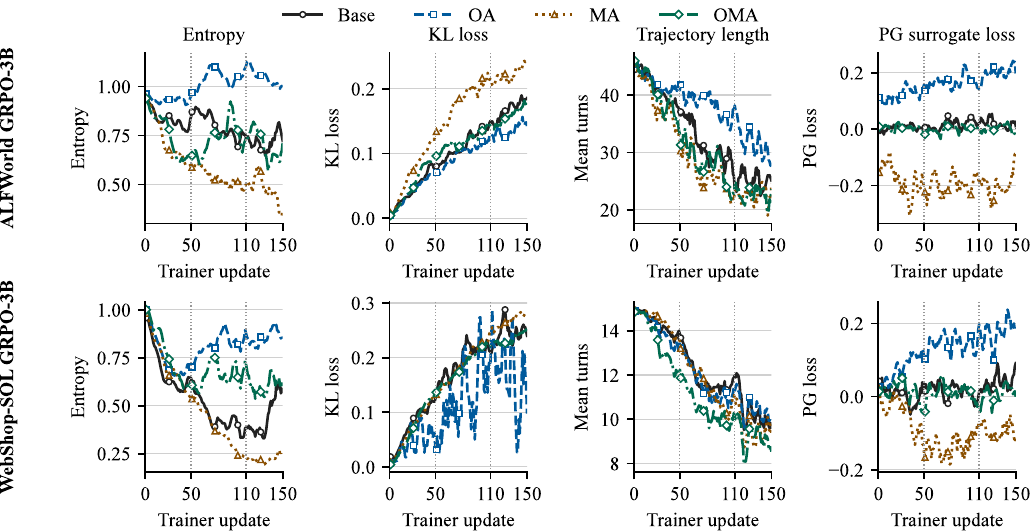}
    \caption{Five-step centered moving averages of actor entropy, KL loss, mean
    episode length, and signed PG loss for the completed ALFWorld-3B and
    WebShop-SOL ablations. The dotted PG reference marks zero; diagnostics
    retain method-native weighting and are interpreted with task performance.}
    \label{fig:supp-dynamics-curves}
\end{figure*}

\section{Complete Stage-Wise and Component Results}
\label{sec:supp-stage-results}

The stage decomposition sharpens the ALFWorld result. At 3B, MA has the largest
early and middle nAUC ($0.388$ and $0.682$), compared with $0.327$ and $0.620$
for Base and $0.346$ and $0.663$ for OMA. In the late window, OMA becomes
strongest ($0.754$), slightly above MA ($0.744$) and Base ($0.741$). The 1.5B
matrix shows the same division: MA is strongest in the early and middle windows
($0.213$ and $0.594$), whereas OMA is strongest in the late window ($0.776$).
Thus Mass Alignment is associated with the clearest early-to-middle
acceleration, while joint alignment produces the strongest late and endpoint
performance.

Ordinary WebShop exhibits a mixed component ordering. OA obtains the strongest
score endpoint ($0.797$), MA the strongest success nAUC ($0.392$), and OMA the
strongest success endpoint ($0.703$); none dominates all four metrics. This
component-level inconsistency matches the weak or reversed expansion--outcome
coupling observed for the benchmark.

The completed WebShop-SOL matrix shows a related but non-identical division of
labor. OA improves success nAUC from $0.399$ to $0.440$, but its endpoint
success is lower than Base ($0.625$ versus $0.648$) and its score metrics are
essentially unchanged. MA improves endpoint success to $0.664$ and obtains the
strongest score nAUC, $0.703$, compared with $0.680$ for Base. OMA is strongest
on strict success, raising endpoint success to $0.734$ and success nAUC to
$0.511$; it also obtains the strongest score endpoint, $0.856$, although its
score nAUC ($0.694$) remains below MA. OMA's success advantage is present in all
three windows: early nAUC increases from $0.137$ to $0.262$, middle from $0.490$
to $0.609$, and late from $0.590$ to $0.674$.

\paragraph{Learning-dynamics diagnostics.}
Figure~\ref{fig:supp-dynamics-curves} compares the full component matrices on
ALFWorld-3B and WebShop-SOL.

On ALFWorld-3B, MA reduces entropy, increases KL, and shortens episodes earliest;
OMA follows less extremely and gives the strongest late performance, while OA
changes these quantities less. The dynamics therefore match MA's early--middle
advantage and OMA's late advantage.

Near the beginning of a PPO update, signed PG loss approximates the negative
objective-weighted mean advantage when the importance ratio is close to one.
Across both settings, OA has a positive mean, MA a negative mean, and OMA a mean
close to zero, while Base has a smaller positive offset. This pattern is
consistent with joint alignment, but clipping and within-update changes also
affect the scalar: zero is not an optimization target, and lower PG loss does not
imply a better policy.

WebShop-SOL shows the same component separation under different outcome
geometry: OA retains higher entropy and lower KL while improving success nAUC;
MA produces stronger displacement and the best score nAUC; OMA has intermediate
entropy/KL dynamics and the strongest strict-success curve. Entropy can reflect
competence or collapse, KL measures displacement rather than quality, and episode
length mixes
behavior with the success/failure composition.

\input{generated/gradient_geometry_table}

\section{Frozen-Batch Gradient Audit}

The audit isolates outer accounting from trajectory generation. For each real
mixed eight-rollout group, it computes one full-model gradient basis vector per
trajectory from mean response log-probability, then combines the unchanged basis
under Raw, OA, MA, and OMA coefficients. The old log-probability is defined as
the detached current value, so the importance ratio is one, clipping is
inactive, and no optimizer step is executed. The available trajectory archive
does not preserve historical token IDs; responses are therefore retokenized
with strict agreement to the recorded token counts.

Table~\ref{tab:gradient-geometry-full} reports the complete paper-facing matrix.
For ALFWorld, increasing failure-to-success response-token asymmetry generally
increases the magnitude of both accounting corrections, especially MA, while
the exact angular relationship remains batch-conditioned. The near-symmetric
WebShop batch keeps all four gradients closely aligned. On the matched
ALFWorld task specification, both GRPO- and GiGPO-trained policies exhibit
strong OA--MA separation, but these rows apply the same ratio-one outer
accounting audit and do not replay GiGPO's native ratio, clipping, or
within-trajectory objective.

As a contract check, the audit also doubles failed-trajectory row multiplicity,
token mass, or both without changing token-level gradient bases. OA remains
invariant to row duplication, MA remains invariant to token-mass inflation, and
OMA remains invariant to both in every reported batch; sensitive negative
controls move in each case. These checks validate the intended accounting
properties, while the main performance and ablation experiments establish
learning outcomes.

\section{Optimizer Instantiations}

OMA replaces only the designated episode-outcome component and the outer
reducer while retaining each optimizer's within-trajectory mechanisms.

\begin{table*}[t]
    \centering
    \footnotesize
    \begin{tabular}{@{}lp{0.24\textwidth}p{0.58\textwidth}@{}}
        \toprule
        Base &
        \multicolumn{1}{l}{\shortstack[l]{Aligned outcome\\component}} &
        \multicolumn{1}{l}{\shortstack[l]{Preserved native\\components}} \\
        \midrule
        GRPO &
        $A_i^{\mathrm{OA}}$ &
        Token ratio and PPO clipping \\
        GiGPO &
        $A_i^{\mathrm{OA}}$ in episode term &
        Anchor grouping, step advantage, and token clipping \\
        GSPO &
        $A_i^{\mathrm{OA}}$ &
        Sequence ratio, detach, and sequence clipping \\
        \bottomrule
    \end{tabular}
    \caption{OMA optimizer interfaces.}
    \label{tab:supp-oma-instantiations}
\end{table*}

\paragraph{GRPO interface.}
Outcome Alignment replaces the episode score used to form the group-relative
advantage and broadcasts the resulting $A_i^{\mathrm{OA}}$ to that trajectory's
valid response tokens. The token likelihood ratio and PPO clipping remain
unchanged. Mass Alignment is applied only after those native tokenwise
surrogates are formed: valid-token contributions are summed within each
trajectory, normalized by that trajectory's valid-token count, and then
averaged across the unique trajectories in the minibatch. OA-only retains the
native token reducer, whereas MA-only retains the native GRPO advantage.

\paragraph{GiGPO interface.}
The designated episode-level term is replaced by
$A_i^{\mathrm{OA}}$, while anchor-state grouping, the row-specific step
advantage, their native mixing, and token-level clipping are preserved. The
combined unreduced row objective is then passed to the same trajectory-balanced
outer reducer.

\paragraph{GSPO interface.}
The aligned trajectory outcome replaces the corresponding episode signal, but
the sequence likelihood ratio, detach construction, and sequence-level clipping
remain native to GSPO. Mass Alignment acts after the clipped sequence objective
is available, equalizing outer trajectory mass rather than changing the
sequence-ratio geometry.

\section{Audit Provenance and Contract Checks}
\label{sec:supp-audit-provenance}

The seven paper-facing gradient rows are fixed in an anonymized scientific audit
record bound by digest \texttt{e43e8937b62a}. The record removes checkpoint
paths, task identifiers, training-repository metadata, and original trajectory
identifiers. Its manifest maps every row to opaque batch metadata, sufficient
combined-gradient Gram entries, file hashes, the ratio-one contract, and
invariance results. Validation verifies every recorded file hash, regenerates
the paper-facing tables, and recomputes the reported token ratios, norm ratios,
cosines, and invariance comparisons.

\newpage
\input{generated/gradient_geometry_provenance}

All audits pass the ratio-one, clipping-inactive contract. Row duplication
leaves OA and OMA invariant, token inflation leaves MA and OMA invariant, and
both perturbations leave only OMA invariant; every sensitive control exceeds
the relative-change threshold.

%% file: generated/supplement_stage_results.tex
\begin{table*}[t]
    \centering
    \footnotesize
    \setlength{\tabcolsep}{5pt}
    \begin{tabular}{@{}llrrrr@{}}
        \toprule
        Setting & Variant & Early & Middle & Late & Endpoint \\
        \midrule
        ALFWorld GRPO (3B) & Base & 32.7 & 62.0 & 74.1 & 74.2 \\
         & OA & 33.6 & 44.0 & 60.2 & 61.7 \\
         & MA & \textbf{38.8} & \textbf{68.2} & 74.4 & 77.3 \\
         & OMA & 34.6 & 66.3 & \textbf{75.4} & \textbf{81.2} \\
        \addlinespace[1pt]
        ALFWorld GRPO (1.5B) & Base & 15.6 & 46.8 & 73.4 & 73.4 \\
         & OA & 11.0 & 22.9 & 35.4 & 38.3 \\
         & MA & \textbf{21.3} & \textbf{59.4} & 75.5 & 76.6 \\
         & OMA & 15.9 & 51.3 & \textbf{77.6} & \textbf{82.8} \\
        \addlinespace[1pt]
        WebShop GRPO (1.5B) & Base & 15.0 & \textbf{45.1} & 60.3 & 65.6 \\
         & OA & 15.1 & 35.4 & 53.5 & 58.6 \\
         & MA & 15.5 & 42.8 & \textbf{63.6} & 52.3 \\
         & OMA & \textbf{15.9} & 42.0 & 63.2 & \textbf{70.3} \\
        \addlinespace[1pt]
        WebShop GRPO (3B) & Base & \textbf{9.7} & 34.2 & \textbf{58.0} & \textbf{64.8} \\
         & OMA & 7.9 & \textbf{42.9} & 54.9 & 57.0 \\
        \addlinespace[1pt]
        WebShop-SOL GRPO (3B) & Base & 13.7 & 49.0 & 59.0 & 64.8 \\
         & OA & 18.4 & 53.2 & 62.1 & 62.5 \\
         & MA & 11.8 & 53.8 & 64.0 & 66.4 \\
         & OMA & \textbf{26.2} & \textbf{60.9} & \textbf{67.4} & \textbf{73.4} \\
        \addlinespace[1pt]
        ALFWorld GiGPO (3B) & Base & \textbf{40.7} & 66.7 & 85.7 & 88.3 \\
         & OMA & 38.8 & \textbf{78.4} & \textbf{86.4} & \textbf{89.8} \\
        \addlinespace[1pt]
        ALFWorld GSPO (3B) & Base & 31.8 & 43.1 & 56.0 & 65.6 \\
         & OMA & \textbf{32.3} & \textbf{53.3} & \textbf{64.2} & \textbf{70.3} \\
        \bottomrule
    \end{tabular}
    \caption*{Complete success-rate stage results (\%). Early, middle, and late are normalized AUC over updates $[0,50]$, $[50,110]$, and $[110,150]$; Endpoint is success at update 150. Bold marks the best value within each setting.}
    \label{tab:supp-stage-results-unnumbered}
\end{table*}

%% file: generated/gradient_geometry_table.tex
\begin{table*}[t]
    \centering
    \small
    \begin{tabular}{llrrrrrr}
        \toprule
        & & & \multicolumn{3}{c}{Gradient-norm ratio to Raw} & & \\
        \cmidrule(lr){4-6}
        Setting & Policy & $r_{\mathrm{tok}}$ & OA & MA & OMA & $\cos(\mathrm{OA},\mathrm{MA})$ & $\cos(\mathrm{Raw},\mathrm{OMA})$ \\
        \midrule
        WebShop & GRPO & 0.99 & 0.94 & 1.02 & 0.96 & 0.996 & 0.996 \\
        ALFWorld & GRPO & 1.14 & 1.19 & 1.26 & 0.94 & 0.261 & 0.998 \\
        ALFWorld & GRPO & 2.17 & 1.93 & 2.72 & 1.30 & -0.421 & 0.947 \\
        ALFWorld & GRPO & 3.59 & 2.41 & 3.46 & 1.31 & -0.506 & 0.985 \\
        ALFWorld & GRPO & 8.78 & 2.44 & 5.05 & 1.60 & -0.368 & 0.982 \\
        \addlinespace[2pt]
        ALFWorld & GRPO & 3.96 & 2.62 & 4.35 & 1.70 & -0.487 & 0.927 \\
        ALFWorld & GiGPO & 3.01 & 2.42 & 3.53 & 1.16 & -0.667 & 0.964 \\
        \bottomrule
    \end{tabular}
    \caption{Frozen-batch full-model gradient geometry. $r_{\mathrm{tok}}$ is the ratio of mean failed-trajectory response tokens to mean successful-trajectory response tokens. The final paired block compares the same ALFWorld task specification under GRPO- and GiGPO-trained policies.}
    \label{tab:gradient-geometry-full}
\end{table*}

%% file: generated/gradient_geometry_provenance.tex
\begin{center}
    \footnotesize
    \setlength{\tabcolsep}{3pt}
    \begin{tabular}{@{}rlllr@{}}
        \toprule
        Row & Audit & Pair & Fingerprint & Inv. \\
        \midrule
        1 & WS base & -- & 0a671241 & 12/12 \\
        2 & ALF low & -- & 422169c1 & 12/12 \\
        3 & ALF median & -- & aca00149 & 12/12 \\
        4 & ALF standard & -- & 07ff99a7 & 12/12 \\
        5 & ALF high & -- & 8e25624f & 12/12 \\
        6 & ALF GRPO pair & a & 4ea26495 & 12/12 \\
        7 & ALF GiGPO pair & a & 21f6a5b4 & 12/12 \\
        \bottomrule
    \end{tabular}
    \captionof{table}{Anonymized gradient-audit snapshot. Fingerprints identify frozen eight-trajectory batches; Inv. reports passed invariance checks.}
    \label{tab:gradient-provenance}
\end{center}